\documentclass{tlp}

\usepackage{amsmath}
\usepackage{graphicx}
\usepackage{multirow}
\usepackage{amsmath,amssymb,amsfonts}
\usepackage{algorithmic}
\usepackage{graphicx}
\usepackage{textcomp}
\usepackage{xcolor}
\usepackage{multirow}
\usepackage{url}
\usepackage{subcaption}
\usepackage[disable]{todonotes}
\usepackage{booktabs}

\begin{document}

\lefttitle{Pennino et al.}

\jnlPage{1}{14}
\jnlDoiYr{2026}
\doival{10.1017/xxxxx}

\title[From Reasoning to Code]{From Reasoning to Code: GRPO Optimization for Underrepresented Languages}
\begin{authgrp}
\author{\sn{Federico} \gn{Pennino}}
\affiliation{Università di Bologna}
\author{\sn{Bianca} \gn{Raimondi}}
\affiliation{Università di Bologna}
\author{\sn{Massimo} \gn{Rondelli}}
\affiliation{Università di Bologna}
\author{\sn{Andrea} \gn{Gurioli}}
\affiliation{Università di Bologna}
\author{\sn{Maurizio} \gn{Gabbrielli}}
\affiliation{Università di Bologna}
\end{authgrp}

\history{\sub{07 02 2026;} \rev{31 03 2026;} \acc{11 05 2026}}

\maketitle

\begin{abstract}
Generating accurate and executable code using Large Language Models (LLMs) remains a significant challenge for underrepresented programming languages, such as Prolog and Lisp, due to the scarcity of public training data compared to high-resource languages like Python. This paper introduces a generalizable Reinforcement Learning (RL) approach that combines small-scale versions of the \texttt{Qwen2.5-Coder} model with Group Relative Policy Optimization (GRPO) to enable effective code generation through reasoning. To address the limitations of sparse datasets, we integrate execution-driven feedback directly into the RL loop, utilizing a reward system that exploits both logical correctness and structural formatting. Experimental results on GSM8K dataset demonstrate significant improvements in reasoning quality and code accuracy across underrepresented languages. These findings underscore the potential of our approach to benefit a wide range of programming languages lacking extensive training resources by leveraging symbolic reasoning and interpreter-based feedback.
\end{abstract}


\section{Introduction}
\label{sec:introduction}

The resurgence of connectionist methods over the past decade has produced Large Language Models of remarkable capability, yet the field of Artificial Intelligence has long drawn on a second tradition: symbolic reasoning, grounded in logic, formal verification, and languages such as Prolog. These two paradigms have historically developed in isolation, and bridging them remains a central challenge for the field. 

From the programming languages perspective this bridge could mean an important (last?) step toward the ``Holy Grail'' of programming introduced by \cite{Freuder1997}: ``The user states the problem and the computer solves it''. This dream has always been a fil rouge of most research in logic programming. We are now very close to making this dream come true, and one promising path is deceptively simple: teach neural models to \emph{write} the symbolic programs themselves. If an LLM can reliably generate executable Prolog code from natural-language specifications, it becomes both a natural-language interface to formal reasoning and a system whose outputs are inspectable, verifiable, and grounded in logic, which are properties that purely neural generation cannot guarantee.

In practice, however, this vision runs into an immediate obstacle. Large Language Models have achieved remarkable proficiency at generating code in mainstream programming languages such as Python and JavaScript, yet this proficiency does not transfer to logic programming. When models such as \texttt{OpenAI Codex}, \texttt{Code-Llama} by \cite{2023arXiv230812950R}, \texttt{StarCoder} by \cite{2023arXiv230506161L}, and \texttt{Qwen2.5-Coder} by \cite{hui2024qwen2} are asked to produce Prolog programs, they consistently fail: they hallucinate non-existent built-in predicates, impose imperative control flow on a declarative paradigm, and mishandle the interplay of unification and backtracking that defines logic programming. The root cause is not architectural but distributional: these languages are scarce in modern pre-training corpora, and no amount of scale will compensate for data that does not exist. 

This paper shows that the obstacle can be addressed by training small, open-weight models with reinforcement learning, using a Prolog interpreter as the source of the signal rather than scarce human-written examples, following the code-evaluation framing of \cite{chen2021codex}.
\begin{figure}
    \centering
    \includegraphics[width=\linewidth]{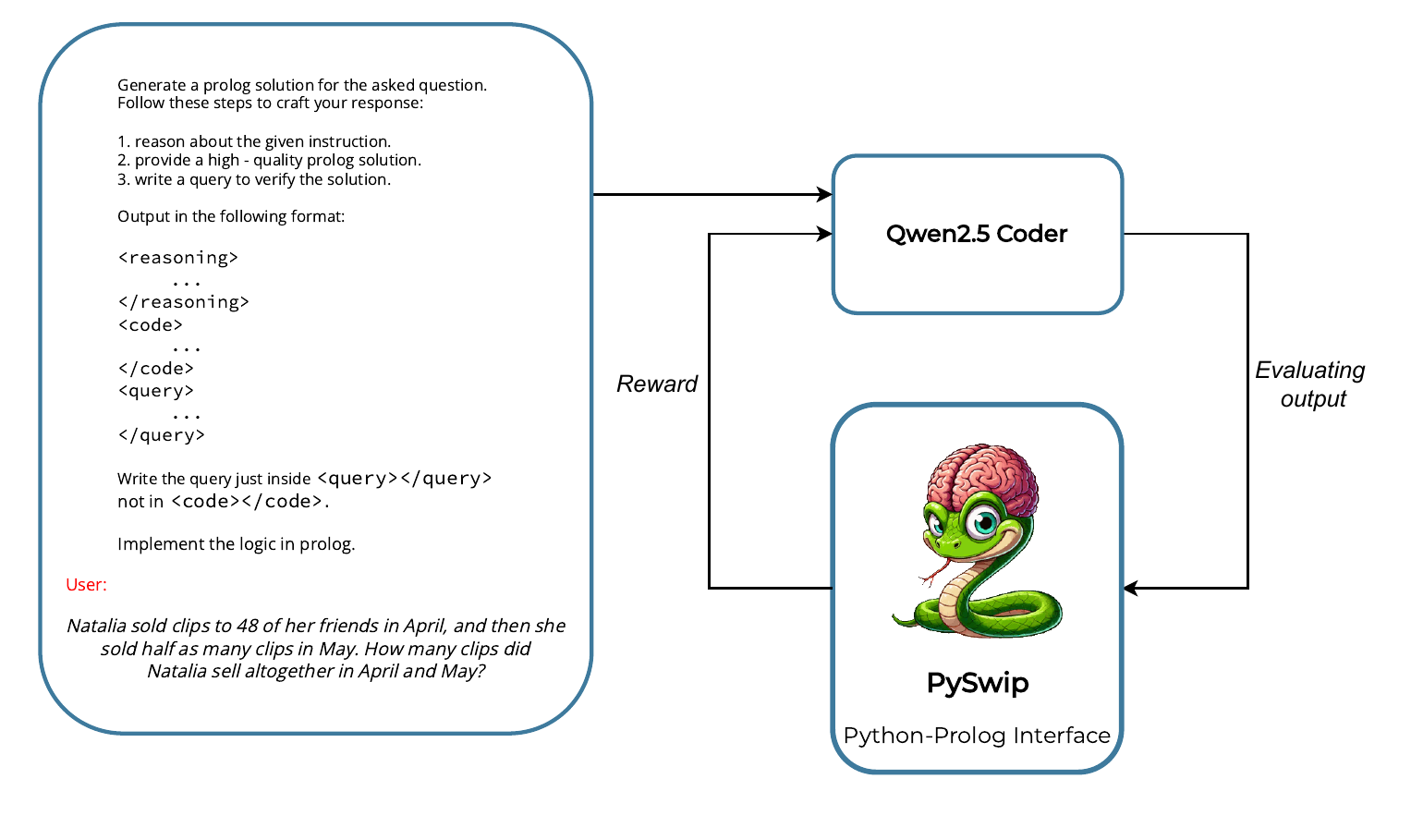}
    \caption{Schematic representation of the GRPO training pipeline. The framework utilizes \texttt{Qwen2.5-Coder} to generate Prolog solutions, which are then evaluated via \texttt{PySwip} to provide execution-based reward signals.}
    \label{fig:training_pipeline}
\end{figure}
More precisely, we explore a methodology that, rather than fine-tuning a model to imitate scarce examples, shifts the learning objective to \emph{maximising execution correctness} as judged by a SWI-Prolog interpreter embedded in the training loop, as shown in \figurename~\ref{fig:training_pipeline}. This is an instance of Reinforcement Learning with Verifiable Rewards (RLVR), discussed by \cite{2025arXiv250614245W}, in which the symbol system serves not as a tool for the neural network but as a teacher that supervises it.

To implement this idea, we adopt Group Relative Policy Optimization (GRPO), introduced by \cite{2024arXiv240203300S}, a policy-gradient algorithm that eliminates the separate value-function network required by Proximal Policy Optimization (PPO).
GRPO estimates the advantage of each generated output relative to a group of peer outputs sampled from the same prompt, reducing computational overhead and stabilising learning in the sparse-reward regime characteristic of code-generation tasks.

We apply GRPO to the Qwen2.5-Coder-Instruct family described by \cite{hui2024qwen2}, scaling from 0.5B to 7B parameters, and train on the GSM8K dataset introduced by \cite{cobbe2021gsm8k}, whose emphasis on structured, logical problem-solving makes it a valuable proxy for the reasoning skills.

First, we identify a failure mode we term \emph{reasoning contraction}: under standard GRPO training with a Kullback--Leibler (KL) divergence penalty, the 7B model collapses to short, memorised output patterns, and its accuracy on GSM8K degrades over training steps (from 0.814 to 0.672 in one-shot).
We attribute this to the KL penalty anchoring the policy too closely to the supervised baseline, thereby impeding the distributional shift required to internalise a fundamentally different programming paradigm.
Second, we show that combining the removal of the KL constraint with a length-aware reward mechanism resolves this contraction: the model's completions expand, its reasoning chains deepen, and its one-shot pass@4 on GSM8K rises to 0.893, thus outperforming both \texttt{PROPER}, introduced by \cite{2024arXiv240517893Y}, a \texttt{Mistral-7B} variant with specialised supervised fine-tuning (0.72), and substantially larger models including \texttt{Llama~3.3 70B} (0.839) and \texttt{Qwen2.5-Coder-32B} (0.873). This work makes four contributions:
\begin{enumerate}
    \item It demonstrates that RLVR with execution-based feedback can compensate for extreme data scarcity, establishing a new state of the art for Prolog code generation on GSM8K with a 7B-parameter model.
    \item It provides a systematic ablation of GRPO training dynamics for logic programming, identifying reasoning contraction as a failure mode of KL-regularised RL and characterising the conditions under which relaxing this constraint is beneficial.
    \item It shows that the resulting gains generalise beyond arithmetic reasoning: improvements transfer to 20 diverse
    Prolog tasks from Rosetta Code and the GSM-Symbolic benchmark, where zero-shot p2 accuracy more than doubles (from 0.298 to 0.673).
    \item It validates the generality of the approach by applying the same framework to Lisp, where zero-shot pass@4 rises from 0.572 to 0.876, suggesting that interpreter-driven RL is effective across distinct families of underrepresented languages.
\end{enumerate}

Taken together, these results demonstrate that for formal languages with a deterministic semantics, the interpreter itself can serve as the primary training signal. This replaces the scarce human data on which current approaches depend, providing a rigorous, automated supervisor that teaches the model to respect the syntax and logic of the target language.

\section{Related work}

Program synthesis, the task of automatically generating executable code from high-level specifications, has traditionally been the domain of symbolic AI. Inductive Logic Programming (ILP), a subfield prominent in the 1990s, used formal logic to infer hypotheses (programs) from positive and negative examples. Systems such as Progol and Aleph can learn recursive logic programs given a strict background knowledge base. However, ILP systems faced severe scalability limitations; they struggled with noise, required extensive domain engineering, and could not process natural language ambiguity.

The deep learning revolution introduced Neural Program Synthesis, replacing the discrete search of ILP with the continuous optimization of neural networks. Early sequence-to-sequence models (e.g., LSTMs) treated code generation as a translation task (Natural Language $\rightarrow$ Code), but struggled with long-range dependencies and the hierarchical structure of abstract syntax trees (ASTs). The Transformer architecture introduced by \cite{vaswani2017attention} and the pre-training paradigm (BERT, GPT) fundamentally altered this landscape.

Modern Code LLMs are decoder-only Transformers trained on vast corpora of source code (e.g., The Stack, GitHub). Despite these advances, the performance of these models is highly uneven. While they excel at Python or Java, their ability to generate Prolog or Lisp is often superficial and merely mimics the syntax (indentation, parentheses) without understanding the semantics (unification, recursion, macro expansion). This is a direct result of the distribution of training data: Prolog code constitutes a vanishingly small percentage of GitHub repositories compared with imperative languages. To address the reliability issues of LLMs in reasoning tasks, researchers have developed frameworks that combine neural intuition with symbolic rigor.

Chain-of-Thought (CoT) prompting encourages LLMs to emit intermediate reasoning steps before the final answer. While this significantly improves performance on math benchmarks such as GSM8K, it suffers from the “validity hallucination” problem: the model can generate a plausible-sounding reasoning chain that contains logical errors, leading to an incorrect answer. Furthermore, CoT relies entirely on the internal representations of the LLM, which are probabilistic and prone to arithmetic drift. 

This led to Program-of-Thought (PoT) prompting, introduced by \cite{chen2022pot}, which advances CoT by decoupling reasoning from computation. In PoT, the LLM generates a program (usually Python) to solve the problem, which is then executed by an external interpreter. This offloads calculation to a deterministic engine, significantly boosting accuracy on arithmetic tasks. 

Recent work has identified Prolog as a superior target for PoT in logical reasoning tasks. GSM8K-Prolog, presented by \cite{2024arXiv240517893Y}, demonstrated that mapping math word problems to Prolog predicates yields higher accuracy than Python PoT for problems requiring constraint satisfaction or symbolic manipulation. Unlike Python, which is imperative (specifying how to calculate), Prolog is declarative (specifying what is true). This aligns more closely with the semantic parsing of natural language logic. The same work also introduced ProPro (Predicate Permutation), a data augmentation method that leverages the commutative property of Prolog rules (i.e., $A, B$ is logically equivalent to $B, A$ in pure Prolog) to improve model robustness.

While SFT effectively teaches syntax, it does not directly optimize for correctness. Reinforcement Learning (RL) allows models to optimize non-differentiable metrics, such as ``does this code compile?'' or ``does this code pass test cases?'' This paradigm leverages the fact that, in domains such as mathematics and programming, the validity of an output can be objectively verified. \cite{le2022coderl} introduced CodeRL, which employed an actor-critic architecture in which unit test results served as the reward signal. RLEF (Reinforcement Learning with Execution Feedback), presented by \cite{rlef2024}, further demonstrated that models can learn to debug their own code by receiving execution traces as feedback, thereby creating an iterative improvement loop. Based on this, we target a setting in which both the training data and the test suites are scarce. By embedding a Prolog interpreter directly into the reward loop, we obtain a dense, ground-truth correctness signal without needing hand-written tests, thereby turning the interpreter from a development tool into a training supervisor.

\section{Methodology}

\subsection{Models and datasets}
This study mainly examined smaller-scale LLMs for Prolog code generation, employing the 0.5B, 1.5B, 3B, and 7B versions of \texttt{Qwen2.5-Coder-Instruct}. These models were chosen for their effectiveness and low resource demands.

A key challenge was the limited availability of publicly accessible Prolog code for supervised training, significantly restricting the models’ initial exposure to Prolog’s syntax. To address this, we adopted a RL strategy using the GSM8K dataset, which, although primarily designed for arithmetic reasoning, its emphasis on structured, logical problem-solving makes it a valuable proxy for evaluating and improving the reasoning skills necessary for Prolog code generation. Training used only the official GSM8K set, and all reported metrics are computed on the official test split to prevent train-test leakage.

\figurename~\ref{fig:training_pipeline} shows an example of the execution pipeline: the LLM learns to map natural language prompts to executable Prolog programs via feedback from execution results.

To evaluate the generalization capabilities of our post-training models, we built a 20-task Prolog dataset sourced from \cite{RosettaCode}, including common algorithmic tasks: Fibonacci sequence, Sieve of Eratosthenes, Quicksort, Binary search, Greatest common divisor, Factorial, Towers of Hanoi, Palindrome detection, Prime decomposition, Dijkstra's Algorithm, Levenshtein distance, N-queens problem, Ackermann function, Balanced brackets, Knight's tour, Merge sort, Roman numerals decode, Longest common subsequence, Huffman coding, 24 game. 
These 20 tasks were selected to represent core logic programming constructs, including recursion, list manipulation, and symbolic backtracking.

We also evaluated models on GSM-Symbolic, a benchmark focused on symbolic reasoning over abstract mathematical expressions. This dataset was created by \cite{gsm-symbolic} solely for testing, with the aim of assessing the generalization capabilities of LLMs trained on GSM8K. We included its variants: GSM-Symbolic-Plus-1 (P1), in which each question includes one additional reasoning clause, and GSM-Symbolic-Plus-2 (P2), with two additional clauses.
These variants increase the complexity of reasoning in a controlled manner: for instance, P2 tasks may involve multi-step comparisons across multiple entities, whereas the base version focuses on single-step calculations.
The GSM-Symbolic setup enables fine-grained analysis of symbolic-reasoning depth, extending beyond GSM8K’s arithmetic focus.
By systematically modifying numerical values and contextual details while preserving the underlying logical structure, we can assess whether the models have learned to extract and encode the core reasoning patterns into Prolog predicates and rules, or whether their translations are overfitted to specific problem templates.

\begin{table}[t]
\centering
\caption{Reward Function Breakdown}
\label{tab:reward_functions}
\resizebox{0.8\linewidth}{!}{%
\begin{tabular}{l|c}
\toprule
\textbf{Reward Function} & \textbf{Reward Range} \\
\midrule
\texttt{xmlcount\_reward\_func} & $[-0.5,0.625]$ \\
\midrule
\texttt{strict\_format\_reward\_func} & $\{0,0.5\}$ \\
\midrule
\texttt{soft\_format\_reward\_func} & $\{0,0.5\}$ \\
\midrule
\texttt{correctness\_reward\_func} & $\{-1,-0.5,1\}$ \\
\bottomrule
\end{tabular}%
}
\end{table}

\subsection{Experimental Setup}
To standardise the training process and ensure reproducible results, we conducted experiments with common customisations. We used TRL, introduced by \cite{vonwerra2022trl}, for the training phase, alongside Low-Rank Adaptation (LoRA), introduced by \cite{2021arXiv210609685H}, with a rank of 32, to enable efficient adaptation and fine-tuning across these configurations.
The evaluation process exploited varying hyperparameters:
\begin{itemize}
    \item{Model size}: Different configurations of the model size were tested, corresponding to 7B, 3B, 1.5B, and 0.5B.
    \item{Checkpoints}: The model was evaluated at three different training checkpoints {500, 1000, 1500}.
    \item{Zero-shot vs. One-shot}: The model was trained in both one-shot and zero-shot learning to measure its ability to generalize based on the context provided. 
\end{itemize}
The resulting models were appropriately prompted based on the learning hyperparameter to be consistent with the prompt setting. The testing script executed inference using the vLLM framework, introduced by \cite{kwon2023efficient}, on a multi-GPU server equipped with 8xNVIDIA A100-SXM4 GPUs (80GB VRAM each), running CUDA 12.2 to ensure efficient execution of large-scale language models. Training and inference were performed on a single GPU unless otherwise specified.

Evaluation metrics and experiment scripts were standardized to facilitate reproducibility and future benchmarking.

\subsection{Reward system}\label{sec:training_pipeline}

The model is trained using GRPO, refining code generation through a multi-objective reward function. The training objective enforces two primary constraints: adherence to the system-prompted template (Fig. \ref{fig:training_pipeline}) and the functional correctness of the generated Prolog code. Total rewards are calculated by aggregating the individual signals summarized in Table \ref{tab:reward_functions}.

\subsubsection{Format reward}
To ensure the Prolog code is extractable and interpretable, we enforce a structured format using \texttt{<reasoning>}, \texttt{<code>}, and \texttt{<query>} tags. This mechanism serves as an early-stage signal.

\begin{itemize}
    \item Tag Presence (\texttt{xmlcount}): Grants +0.125 for each required tag. A penalty of -0.5 is applied if the query is incorrectly nested inside the \texttt{<code>} block to prevent execution errors;
    \item Strict Evaluation: A binary +0.5 reward if the output matches the template exactly;
    \item Soft Evaluation: A binary +0.5 reward if the code is extractable via regex, even if minor formatting errors exist outside the tags.
\end{itemize}

This structured formatting strategy increases sample efficiency by penalizing structural deviations, allowing the model to quickly converge on a consistent and executable output format.

\subsubsection{Correctness reward}

We evaluate semantic correctness by executing the generated code via PySwip, a Python interface for SWI-Prolog. The environment is sandboxed with a timeout to prevent infinite loops. The reward is assigned based on the interpreter’s output compared to the ground truth:

\begin{figure*}[t!]
  \centering
  \begin{subfigure}{0.49\linewidth}
    \centering
    \includegraphics[width=\linewidth]{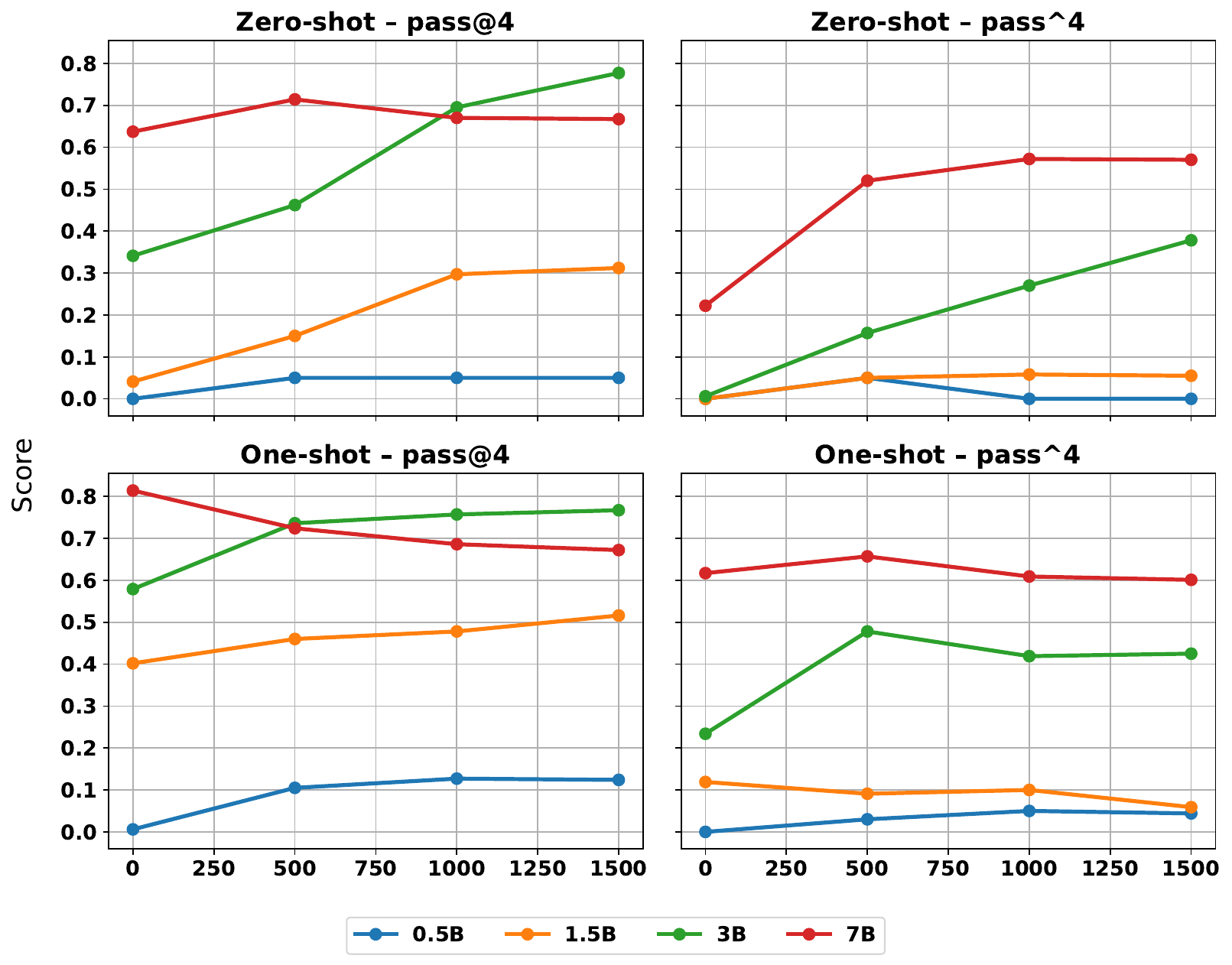}
    \caption{GSM8K}
    \label{fig:gsm8k}
  \end{subfigure}
  \hfill
  \begin{subfigure}{0.49\linewidth}
    \centering
    \includegraphics[width=\linewidth]{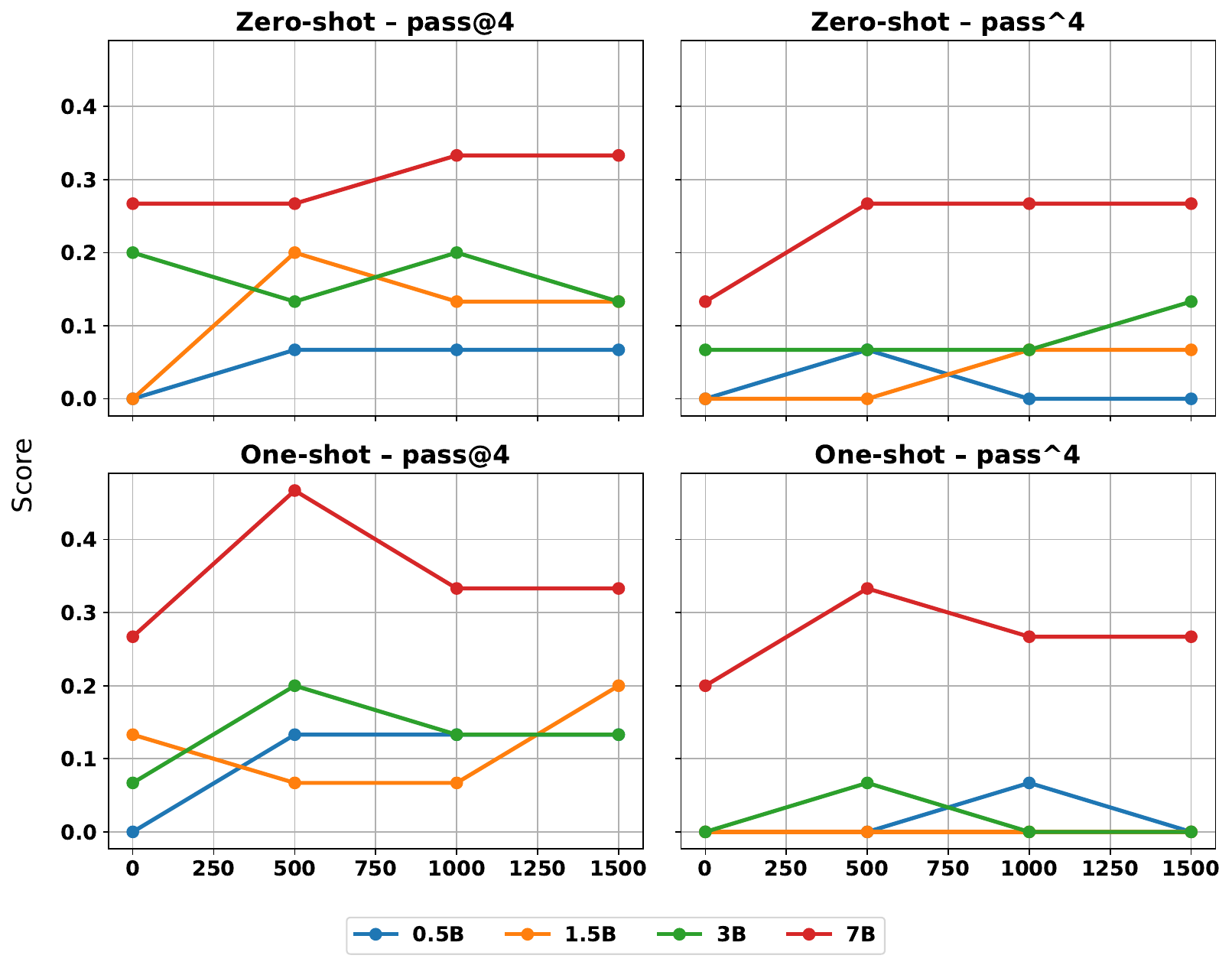}
    \caption{Rosetta Code}
    \label{fig:rosetta}
  \end{subfigure}
  \caption{
    pass@4 and pass$^\wedge 4$ results for \texttt{Qwen2.5-Coder} models in zero-shot and one-shot settings.
  }
  \label{fig:rosetta_gsm8k}
\end{figure*}

\begin{itemize}
    \item Success (+1.0): The execution result matches the expected answer exactly;
    \item Logical Error (-1.0): The code executes but returns an incorrect result;
    \item Syntax Error (-0.5): The code is malformed and fails to compile/execute;
    \item Timeout/No Output (-0.1): The interpreter fails to return a result within the limit of 5 seconds.
\end{itemize}

By penalizing syntax errors more heavily than timeouts, we prioritize generating “runnable” code during the initial phases of training.

\subsection{Evaluation metrics}\label{sec:metrics}
We assess the model’s capability to generate functionally correct Prolog code through evaluation. For each problem instance in our dataset, the model generates $k=4$ candidate solutions. We employ two primary metrics based on this execution framework to quantify performance: Pass$@k$ and Pass$^\wedge k$.

\textbf{Pass$@k$} This metric measures the success rate considering the $k$ generated candidates per problem. A problem is deemed successfully solved if \textit{at least one} of the $k$ generated solutions produces the correct output when executed. It is calculated over the entire dataset of $N$ problems as shown in Eq.~\ref{eq:top_4_any_accuracy}. In this formula, $N$ is the total number of problems, $c_i^j$ is the $j$-th code for the $i$-th problem, $P(c_i^j)$ is its execution result via interpreter $P$, $r_i$ is the ground truth, and the disjunction ($\bigvee$) captures the “at least one” success condition.
\begin{align}\label{eq:top_4_any_accuracy}
pass@k = \frac{1}{N}\sum_{i=1}^{N} \bigg(\bigvee_{j=1}^{k}P(c_i^j)=r_i\bigg)
\end{align}

\textbf{Pass$^\wedge k$} Providing a stricter assessment of performance, this metric requires \textit{all} $k$ generated solutions for a given problem to execute correctly. It measures the model’s consistency and reliability in generating valid code. The formal definition is shown in Eq~\ref{eq:top_4_all_accuracy}.
\begin{align}\label{eq:top_4_all_accuracy}
pass^\wedge k= \frac{1}{N}\sum_{i=1}^{N}\bigg(\bigwedge_{j=1}^{k}P(c_i^j)=r_i\bigg)
\end{align}
The notation follows Equation \ref{eq:top_4_any_accuracy}, but the conjunction ($\bigwedge$) enforces the requirement that all $k$ generated candidates must match the ground truth $r_i$ for the problem to be counted as correct. This stricter metric evaluates the model’s reliability across multiple generations of attempts.

\begin{figure}
    \centering
    \includegraphics[width=\linewidth]{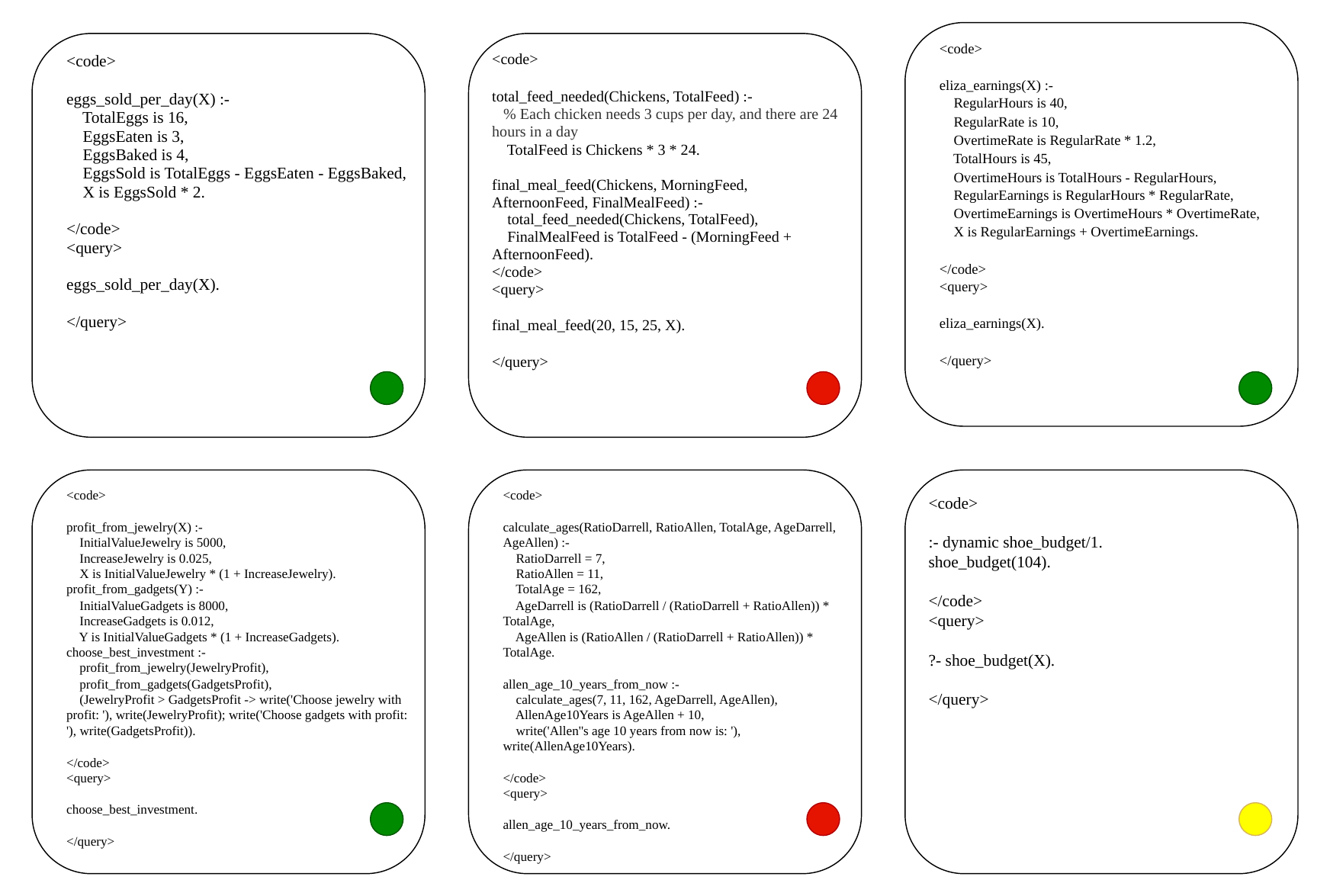}
    \caption{Examples of Prolog code generated by the trained models on GSM8K problems. Green dots indicate solutions that executed correctly and returned the expected answer. Red dots denote solutions that failed due to logical or syntax errors. Yellow dots highlight generations that, while syntactically plausible, were just hardcoding the final value.}
    \label{fig:prolog_examples}
\end{figure}

\begin{table}[t]
         \centering
         \caption{pass@4 scores for \texttt{Qwen2.5-Coder} trained with new reward function, reporting results for both standard and length-optimized reward functions.}
         \label{tab:length_reward}
         \resizebox{\linewidth}{!}{%
         \begin{tabular}{l|l|cccccc}
         \toprule
         \textbf{Configuration} & $\quad$ \textbf{Model} & \textbf{Base} & \textbf{500} & \textbf{1000} & \textbf{1500} \\
         \midrule
        \cite{2024arXiv240517893Y}  & \texttt{Mistral 7B} & \textbf{0.72}  & $-$  & $-$  & $-$ \\
         \midrule
         Zero-shot  & \texttt{Llama 3.3 70B}  & \textbf{0.839}  & $-$  & $-$  & $-$ \\
         \midrule
         Zero-shot  & \texttt{Qwen2.5-Coder-32B} & \textbf{0.873}  & $-$  & $-$  & $-$ \\
         \midrule
         Zero-shot  & $\quad$ \texttt{Qwen2.5-Coder-7B}        & 0.637  & \textbf{0.714}  & 0.670  & 0.667  \\
                    & $\quad$  + length constraint &       & 0.764  & \textbf{0.782}  & 0.780  \\
                    & $\quad$  + no KL divergence  &       & 0.847  & 0.880  & \textbf{0.882}  \\
         \midrule
         One-shot   & $\quad$ \texttt{Qwen2.5-Coder-7B}        & \textbf{0.814}  & 0.724  & 0.686  & 0.672  \\
                    & $\quad$ + length constraint &       & 0.852  & \textbf{0.862}  & 0.856  \\
                    & $\quad$  + no KL divergence  &       & 0.879  & 0.883  & \textbf{0.893}  \\
         
         \bottomrule
         \end{tabular}%
         }
     \end{table}
\section{Results and Discussion}
\label{Section_results}

\subsection{GSM8K Dataset}
The performance of \texttt{Qwen2.5-Coder} models on the GSM8K benchmark reveals distinct behaviour in logic programming with respect to model size. As illustrated in Fig. \ref{fig:rosetta_gsm8k}a, the smallest variant, \texttt{Qwen2.5-Coder-0.5B}, is initially not able to learn in zero-shot settings with a $0.00$ accuracy, highlighting a fundamental lack of Prolog-specific structural priors. However, introducing a single demonstration (one-shot) serves as a \textit{syntax anchor}, enabling the model to reach $0.13$ accuracy after $1000$ steps.

As model size scales to 3B and 7B, we observe the emergence of zero-shot generalization capabilities. Notably, while the 7B model achieves a high one-shot baseline of $0.81$, its performance is worsening during training, settling at $0.67$ after 1500 steps. Instead, the 3B model significantly improves its performance during training, from 0.35 to 0.78. The 7B “performance drift” suggests that without specific GRPO fine-tuning, RL may lead the model to overlook strict prolog syntax, a phenomenon further evidenced by the decline in the strict $pass^\wedge4$ metric over training steps.

\figurename~\ref{fig:prolog_examples} illustrates representative Prolog solutions produced across different training configurations. Correctly executing solutions (green) demonstrate the model’s ability to translate arithmetic word problems into well-structured predicates with appropriate variable bindings and queries. Failed generations (red) typically stem from mishandled predicate arity or incorrect arithmetic decomposition. The yellow dot marks a generation that, while syntactically valid, hardcodes intermediate values directly into the predicate body rather than encoding general-purpose logic — a pattern frequently observed among smaller model variants. 

\subsection{Rosetta Code}
To evaluate if reasoning gains transfer to general programming, we tested the models on Rosetta Code (Fig. \ref{fig:rosetta_gsm8k}b). The results confirm that GRPO adaptation is efficient for low-resource languages: the 1.5B model, which was non-functional at base, reached a $pass@4$ of $0.20$ within just $500$ steps; The 7B model peaked early at $500$ steps ($0.46$ one-shot), nearly doubling its zero-shot baseline.
The persistent gap between zero-shot and one-shot performance suggests that even after reasoning-based fine-tuning, example-based prompting remains essential for maintaining correct predicate arity and recursion patterns in traditionally underrepresented languages.

\section{Ablation Study}

\begin{figure*}[b!]
  \centering
  \begin{subfigure}{0.49\linewidth}
    \centering
    \includegraphics[width=\linewidth]{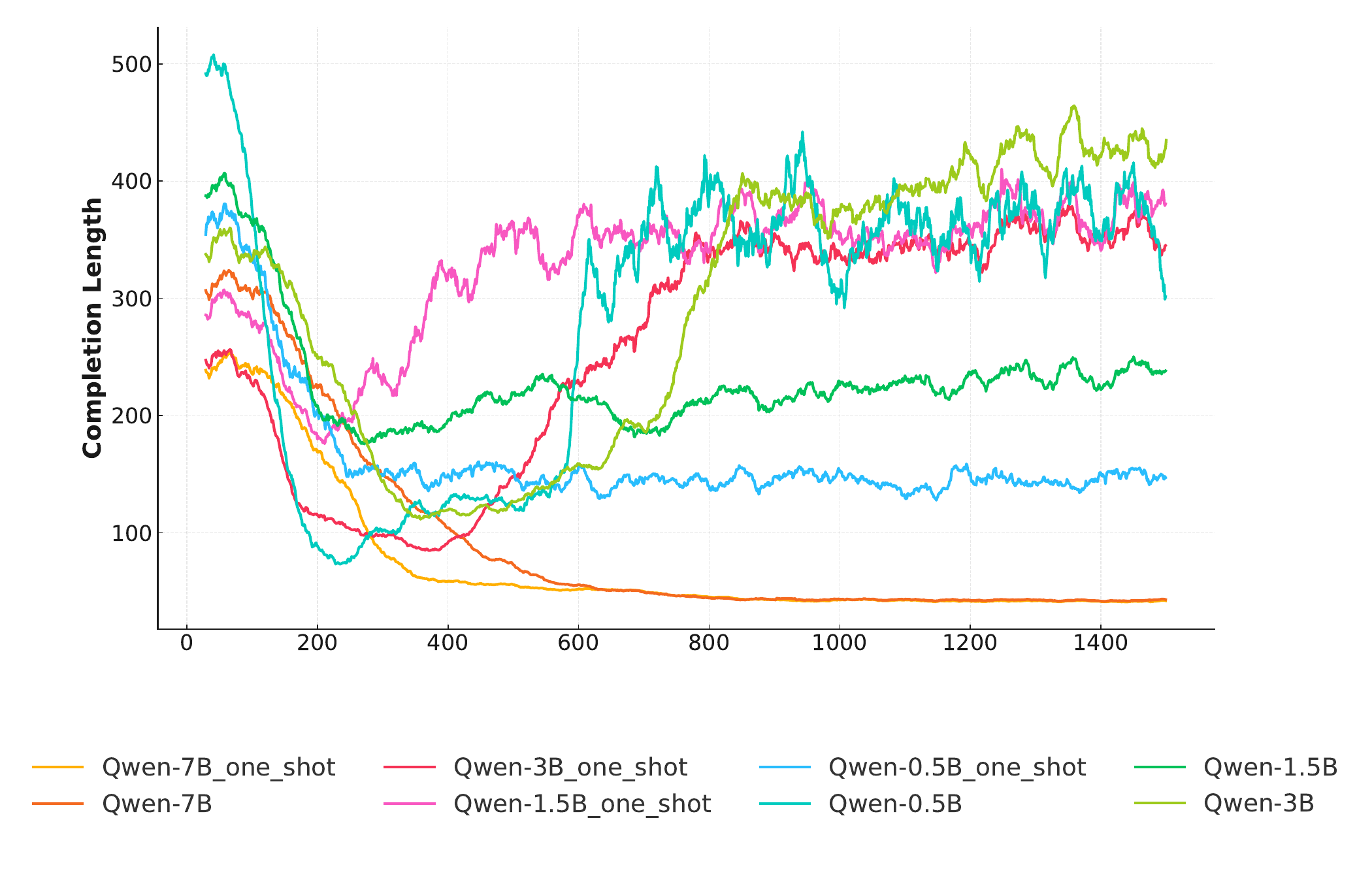}
    \caption{GSM8K}
    \label{fig:clvstraining}
  \end{subfigure}
  \hfill
  \begin{subfigure}{0.49\linewidth}
    \centering
    \includegraphics[width=\linewidth]{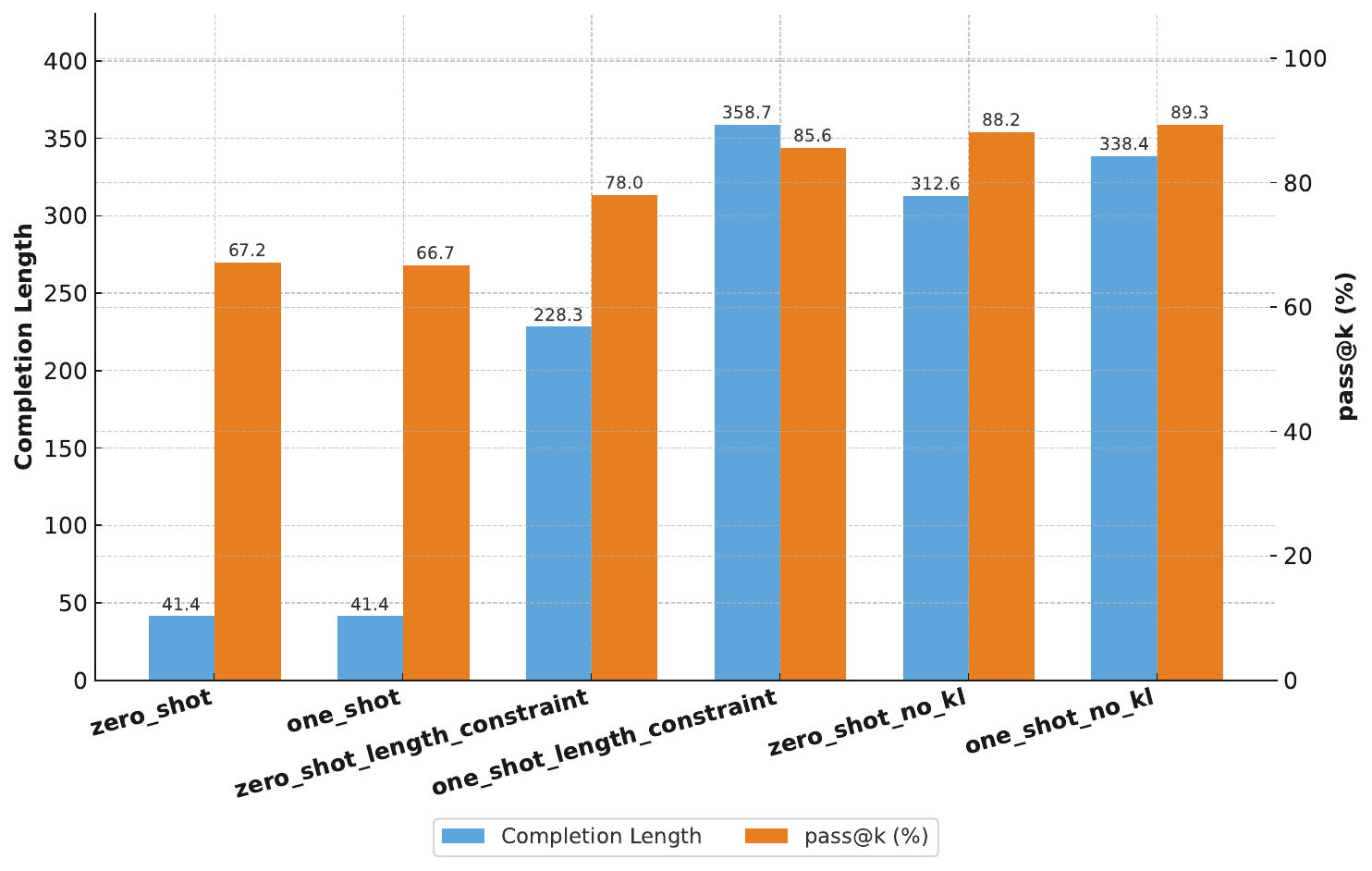}
    \caption{Rosetta Code}
    \label{fig:comparision_type}
  \end{subfigure}
  \caption{
    (a) Completion length for six Qwen model variants (0.5B, 1.5B, 3B, 7B with and without one-shot prompting) over training steps; (b) Comparison of different strategies applied to \texttt{Qwen2.5-Coder-7B}: Completion Length (left y-axis) and pass@k (right y-axis). Models are grouped by level of supervision (zero-shot vs. one-shot) and whether length constraints or KL-regularization are applied.
  }
  \label{fig:length_and_model_comparison}
\end{figure*}

\begin{table}[t]
    \centering
    \caption{pass@4 scores for GSM-Symbolic p1 and p2 datasets}
    \label{tab:length_reward_gsmsym_combined}
    \resizebox{\linewidth}{!}{%
    \begin{tabular}{l|l|l|cccc}
        \toprule
        \textbf{Dataset} & \textbf{Configuration} & \quad \textbf{Model} & \textbf{Base} & \textbf{500} & \textbf{1000} & \textbf{1500} \\
        \midrule
        \multirow{6}{*}{\textbf{p1}} & Zero-shot & \quad \texttt{Qwen2.5-Coder-7B} & 0.469 & \textbf{0.535} & 0.418 & 0.427 \\
        & & \quad + length constraint & & 0.575 & \textbf{0.638} & 0.634 \\
        & & \quad + no KL divergence & & 0.679 & \textbf{0.763} & 0.752 \\
        \cmidrule{2-7}
        & One-shot & \quad \texttt{Qwen2.5-Coder-7B} & \textbf{0.754} & 0.578 & 0.479 & 0.470 \\
        & & \quad + length constraint & & 0.747 & 0.782 & \textbf{0.787} \\
        & & \quad + no KL divergence & & 0.770 & 0.792 & \textbf{0.803} \\
        \midrule
        \midrule
        \multirow{6}{*}{\textbf{p2}} & Zero-shot & \quad \texttt{Qwen2.5-Coder-7B} & 0.298 & \textbf{0.336} & 0.197 & 0.198 \\
        & & \quad + length constraint & & 0.388 & \textbf{0.451} & 0.433 \\
        & & \quad + no KL divergence & & 0.534 & 0.644 & \textbf{0.673} \\
        \cmidrule{2-7}
        & One-shot & \quad \texttt{Qwen2.5-Coder-7B} & \textbf{0.588} & 0.398 & 0.205 & 0.201 \\
        & & \quad + length constraint & & 0.645 & 0.665 & \textbf{0.679} \\
        & & \quad + no KL divergence & & 0.673 & \textbf{0.721} & 0.717 \\
        \bottomrule
    \end{tabular}
    }
\end{table}
\subsection{Prolog Code Generation}

As visualized in \figurename~\ref{fig:clvstraining}, there is a correlation between the expansion of completion length and the final code accuracy; specifically, the transition from short reasoning chains to longer length sequences is the primary driver of the observed performance leaps. During initial training, we observed a “reasoning contraction” in which the 7B model shortened its output, at the cost of expressivity. 

To counteract this effect, we introduce a length-constrained reward on the reasoning segment. Given a completion $( c )$, let ( $r_\text{length}(c)$ ) be the token-length of the reasoning (scaled by $(10^{-4})$). We define the length constraint reward as: 
\[ 
r(c) =
\begin{cases}
1 & \text{if } 0.009 < r_{\text{length}}(c) < 0.013 ; \\
0 & \text{otherwise}
\end{cases}
\]
This formulation encourages reasoning traces within a target length range.

By implementing a length-aware reward mechanism and removing the KL-divergence penalty, we effectively expanded the model’s planning process. As shown in Table \ref{tab:length_reward}, these modifications shifted the zero-shot $pass@4$ from $0.637$ to a peak of \textbf{$0.882$} at $1500$ steps. The impact of these modifications is summarized in \figurename~\ref{fig:comparision_type}, which illustrates the concurrent increase in both completion length and pass@k performance across zero-shot and one-shot settings.

According to the results in Table \ref{tab:length_reward}, our optimized \texttt{Qwen2.5-Coder-7B} configuration --- utilizing length constraints and the removal of KL divergence --- achieves a peak one-shot score of 0.893, remarkably outperforming \texttt{PROPER} (0.72), a Mistral 7B variant that underwent specialized supervised fine-tuning for Prolog code generation in GSM8K. Furthermore, our model significantly outperforms larger general-purpose architectures, such as \texttt{Llama 3.3 70B} (0.839) and \texttt{Qwen2.5-Coder-32B} (0.873), on the same benchmark. Baselines for \texttt{Llama 3.3 70B} and \texttt{Qwen2.5-Coder 32B} were evaluated using the same system prompts and XML-based output templates to ensure a fair comparison.

\subsection{GSM-Symbolic}
To ensure the model’s reasoning capabilities are truly robust and not overfitted to the standard GSM8K templates, we evaluated our configurations on GSM-Symbolic (p1 and p2), as detailed in Table \ref{tab:length_reward_gsmsym_combined}. A critical observation from this evaluation is the significant performance degradation caused by standard GRPO fine-tuning on reasoning-heavy tasks; specifically, in the one-shot setting on the p2 dataset, accuracy dropped from a baseline of $0.588$ to $0.201$ after $1500$ steps.

Our length-aware reward mechanism effectively mitigated this decline, while removing the KL-divergence penalty further improved performance on both benchmarks. On the p1 dataset, the one-shot accuracy rose from $0.754$ to $0.803$. More notably, on the more demanding p2 dataset—which requires deeper symbolic manipulation—the zero-shot score more than doubled, increasing from $0.298$ to \textbf{$0.673$}. These results demonstrate that our methodology enhances genuine symbolic reasoning and cross-task robustness, ensuring that the model can handle structural variations in logic problems rather than relying on simple memorization of patterns.

\subsection{Lisp Code Generation}

To validate the versatility of our approach, we also evaluated Lisp (Table \ref{tab:lisp}). The base \texttt{Qwen2.5-Coder-7B} model exhibited higher initial proficiency in Lisp ($0.5716$) than in Prolog, likely due to a stronger representation in its pretraining data. Our adaptation significantly improved performance, reaching a peak zero-shot score of \textbf{$0.8757$}. 

However, the impact of removing the KL-divergence was notably more modest in Lisp ($0.8712$ to $0.8757$) than the transformative gains observed in Prolog. This suggests that “un-regularized” training is most critical for highly specialized, low-resource languages such as Prolog, where the model must deviate substantially from its base policy to achieve formal correctness. For Lisp, once reasoning length is optimized, policy regularization is less of a limiting factor.

\begin{table}[t]
    \centering
    \caption{pass@4 scores for GSM-8K dataset for Lisp}
    \label{tab:lisp}
    \resizebox{\linewidth}{!}{%
    \begin{tabular}{l|l|cccccc}
        \toprule
        \textbf{Configuration} & $\quad$ \textbf{Model} & \textbf{Base} & \textbf{500} & \textbf{1000} & \textbf{1500} \\
        \midrule
        Zero-shot  & $\quad$ \texttt{Qwen2.5-Coder-7B}                           & 0.5716  & 0.830  & 0.8653  & \textbf{0.8702}    \\
                   & $\quad$ + length constraint &      & 0.837  & \textbf{0.8712}  & 0.8702  \\
                   & $\quad$ + no KL divergence &      & 0.840  & \textbf{0.8757}  & 0.8711  \\
        \midrule
        One-shot   & $\quad$ \texttt{Qwen2.5-Coder-7B}                           & 0.8112  & 0.851     & \textbf{0.868}      & 0.863      \\
                   & $\quad$ + length constraint &      & 0.851  & \textbf{ 0.871}  & 0.864  \\
                   & $\quad$ + no KL divergence &      & 0.861  & \textbf{0.873}  & 0.868  \\
        \bottomrule
    \end{tabular}%
    }
\end{table}

\section{Conclusion}

This work validates GRPO as an effective method for improving Prolog code generation in \texttt{Qwen2.5-Coder}. We observe clear gains across sizes, here reported as pass@4 scores and absolute deltas: the 0.5B model improves by +0.118 in one-shot, the 1.5B model reaches 0.312 zero-shot, the 3B model gains +0.436 zero-shot, and the 7B model without KL reaches 0.882 (zero-shot) and 0.893 (one-shot), establishing a new state of the art on Prolog GSM8K under our evaluation protocol. Improvements generalize to broader programming tasks (Rosetta Code: 7B one-shot +0.200) and transfer to Lisp (zero-shot +0.299, from 0.5716 to 0.8702; peaking at 0.8757 without KL; one-shot 0.873 without KL). KL removal helps most in Prolog, and length-aware rewards provide consistent additional gains. These results indicate a robust, adaptable path for enhancing code generation in underrepresented languages that often lack extensive training datasets and could benefit significantly from our methodology.

Future work could investigate hybrid neural-symbolic architectures where LLMs delegate complex logical inference or constraint satisfaction to a Prolog or Lisp engine, moving beyond simple code generation. Extending program-of-thought approaches, LLMs could learn to generate intermediate Prolog queries to structure their reasoning. Furthermore, Prolog’s capabilities can be leveraged for more sophisticated verification, checking the logical consistency of LLM reasoning steps or formal properties of generated code, potentially using execution traces for enhanced explainability. Moreover, given the success demonstrated with Prolog and Lisp, we plan to extend our study to other underrepresented languages.

Finally, we are considering the construction of a new training dataset specifically designed to better capture non-numerical reasoning patterns, potentially addressing limitations of GSM8K in this regard. Such a dataset could enable more robust evaluation and training of models on symbolic and logical reasoning tasks beyond arithmetic problem solving.

\section{Data Availability}

The source code for preprocessing and evaluation is at:

\url{https://github.com/biancaraimondi/LLM_Format}

\url{https://github.com/biancaraimondi/LLM_Format}

\end{document}